\documentclass[conference]{IEEEtran}
\IEEEoverridecommandlockouts
\usepackage{cite}
\usepackage{amsmath,amssymb,amsfonts}
\usepackage{algorithmic}
\usepackage{graphicx}
\usepackage{textcomp}
\usepackage{xcolor}
\def\BibTeX{{\rm B\kern-.05em{\sc i\kern-.025em b}\kern-.08em
    T\kern-.1667em\lower.7ex\hbox{E}\kern-.125emX}}

\usepackage{cite}  
\usepackage{caption}
\usepackage{subcaption}
\usepackage{multirow}
\usepackage{enumitem}
\usepackage{amssymb,amsfonts}
\usepackage{algorithmic}
\usepackage{textcomp}
\usepackage{xcolor}
\usepackage{hyperref}
\usepackage{tikz}
\usetikzlibrary{positioning}

\newcommand{\Ls}{\mathcal{L}}

\DeclareMathOperator*{\argmax}{arg\,max}

\title{Maximum Likelihood Distillation for Robust Modulation Classification
\thanks{This work has been sponsored by armasuisse Science and Technology under the project ARNO (project code ARAMIS 047-22).}
}

\author{\IEEEauthorblockN{Javier Maroto}
\IEEEauthorblockA{\textit{Signal Processing Laboratory (LTS4)} \\ \textit{EPFL, Switzerland}}
\and
\IEEEauthorblockN{Gérôme Bovet}
\IEEEauthorblockA{\textit{armasuisse Science\&Technology} \\
\textit{Cyber-Defence Campus, Switzerland}}
\and
\IEEEauthorblockN{Pascal Frossard}
\IEEEauthorblockA{\textit{Signal Processing Laboratory (LTS4)} \\ \textit{EPFL, Switzerland}}}

\begin{document}
\maketitle

\begin{abstract}
Deep Neural Networks are being extensively used in communication systems and Automatic Modulation Classification (AMC) in particular. However, they are very susceptible to small adversarial perturbations that are carefully crafted to change the network decision. In this work, we build on knowledge distillation ideas and adversarial training in order to build more robust AMC systems. We first outline the importance of the quality of the training data in terms of accuracy and robustness of the model. We then propose to use the Maximum Likelihood function, which could solve the AMC problem in offline settings, to generate better training labels. Those labels teach the model to be uncertain in challenging conditions, which permits to increase the accuracy, as well as the robustness of the model when combined with adversarial training. Interestingly, we observe that this increase in performance transfers to online settings, where the Maximum Likelihood function cannot be used in practice. Overall, this work highlights the potential of learning to be uncertain in difficult scenarios, compared to directly removing label noise.
\end{abstract}

\begin{IEEEkeywords}
Neural networks, Robustness, Maximum Likelihood, knowledge distillation, automatic modulation classification
\end{IEEEkeywords}

\section{Introduction}

Communication systems often require the receiver to recognize the modulation scheme that has been used to encode the transmitted signals. The task is often referred to as Automatic Modulation Classification (AMC) and has applications ranging from detecting daily radio stations and managing spectrum resources, to eavesdropping and interfering with radio communications. The AMC task can be formulated as a Maximum likelihood (ML) estimation problem \cite{huan1995likelihood,dobre2007survey,hameed2009likelihood}, by computing the maximum of the likelihood function of the received signal with all possible modulations. Even if this solution is Bayes-optimal, sub-optimal approximations are used in practice \cite{dobre2007survey,hameed2009likelihood} because ML requires prior knowledge of the channel characteristics and is computationally complex to solve in realtime.

Deep learning \cite{goodfellow2016deep} has been proposed as a better solution for AMC as neural networks are relatively fast at inference time, adaptable to any channel characteristics, and their performance scales well with large quantities of data. Two classes of neural network architectures have been proposed for this task: long short-term memory networks (LSTMs) \cite{Rajendran_Meert_Giustiniano_Lenders_Pollin_2018,Guo_Jiang_Wu_Zhou_2020} and convolutional neural networks (CNNs) \cite{OShea_Corgan_Clancy_2016,West_OShea_2017,Sadeghi_Larsson_2019}. In particular, the authors in \cite{OShea_Roy_Clancy_2018} use a model based on the ResNet architecture \cite{Szegedy_Ioffe_Vanhoucke_Alemi_2016}, that outperforms all other CNN architectures. Other works use neural architecture search (NAS) \cite{ahmadi2008modulation,dai2019multi,perenda2021evolutionary} to trade off performance with computational speed. Even though neural networks currently represent the main and most effective method to tackle the AMC problem \cite{OShea_Roy_Clancy_2018}, they are very susceptible to adversarial perturbations, which compromises the security of the system against malicious attacks  \cite{szegedy2014intriguing, moosavi2017universal,Sadeghi_Larsson_2019,lin2020threats,Flowers_Buehrer_Headley_2019, maroto2021benefits}. 

Adversarial perturbations are carefully crafted but almost imperceptible disturbances that are added to the transmitted signal. These adversarial perturbations can be used as an effective way of jamming wireless communication systems as they require much less energy than normal attacks. Algorithms like FGSM \cite{goodfellow2014explaining} or PGD \cite{Madry_Makelov_Schmidt_Tsipras_Vladu_2019} can compute such perturbations by solving the following optimization problem:
\begin{equation}
\label{eq:adv_pert}
    \delta_i^* = \argmax_{\delta_i}\Ls(f_{\theta}(x_i + \delta_i), y_i) \quad \text{s.t.} \quad \lVert \delta_i \rVert_{\infty} \leq \varepsilon
\end{equation}
where $\delta_i^*$ is the adversarial perturbation added to the received signal $x_i$, $y_i$ is the one-hot-encoded vector of length K that encodes the modulation used, $\Ls$ is the cross-entropy loss, $f_{\theta}$ is the neural network defined with weights $\theta$, and $\varepsilon$ is a fixed value that constrains the norm of the perturbation to be small and imperceptible. On the other hand, adversarial training \cite{maroto2021safeamc,manoj2022toward}, randomized smoothing \cite{kim2021channel,manoj2022toward} and Generative Adversarial Networks (GANs) \cite{wang2022gan} have been shown to be effective in increasing the robustness of AMC models. But, in addition to being computationally costly, these defenses greatly reduce the accuracy of the model on clean data.

Some works from computer vision have tried to understand why networks are so susceptible to attacks, and label noise has been pointed to as one of the main causes \cite{sanyal2020benign}. Knowledge distillation~\cite{hinton2015distilling,romero2014fitnets,zagoruyko2016paying,chebotar2016distilling} can reduce this effect by training a student model to match the outputs of a teacher model. While knowledge distillation has been used in AMC~\cite{ma2020cross}, some computer vision works~\cite{goldblum2020adversarially,zi2021revisiting,shao2021and,maroto2022benefits} have shown that to achieve adversarial robustness, the outputs of the student model have to be matched on adversarial examples as well. And, to achieve a significant increase in robustness, both the teacher and the student models have to be trained adversarially, which can be very costly computationally.

In this work, we build on the above observations and propose a new data-centric method to make neural networks more accurate and robust to adversarial attacks. We make the hypothesis that the information given by the original labels is limited, and does not encode the difference in uncertainty between low and high SNR signals. 
Therefore, we propose to bring better information to the model by distilling the theoretical class probabilities derived from the offline maximum likelihood function estimation. We train a neural network on these computed probabilities to transfer the better performance of the maximum likelihood function, and then use the resulting network for inference in online settings. Thus, not requiring priors only available in offline settings. Moreover, we believe our proposal is especially interesting when training robust models. Because adversarial training performance is quite dependent on the quality of the data, it benefits from the higher quality information given by these distilled probabilities.

We show experimentally that such knowledge distillation permits to achieve better accuracy and adversarial robustness, compared with using the original labels or removing label noise. 
Then, we show similar increases in performance when we combine this approach with adversarial training, where we train the model to match the distilled labels on adversarial examples. We highlight, that despite maximum likelihood being limited to offline settings, we see a clear benefit in the resulting model even in online settings. We claim that helping the model be uncertain in challenging scenarios is a promising direction with relevance outside of AMC settings, further validating the intuition that the choice and the quality of the training data are highly influential on the adversarial robustness of the model.

\section{Framework}

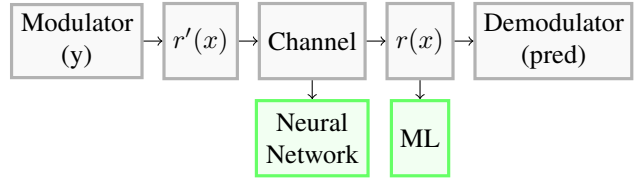
\begin{figure}
\begin{tikzpicture}[
	greennode/.style={rectangle, draw=green!60, fill=green!5, very thick, minimum size=5mm},
	graynode/.style={rectangle, draw=gray!60, fill=gray!5, very thick, minimum size=5mm},
	]
	\node[graynode,text width=1.5cm,text centered](mod){Modulator\\(y)};
	\node[graynode](tf)[minimum height = 1cm][text centered][right=0.25cm of mod]{$r'(x)$};
	\node[graynode](channel)[minimum height = 1cm, right=0.25cm of tf]{Channel};
	\node[graynode](rf)[minimum height = 1cm, right=0.25cm of channel]{$r(x)$};
	\node[graynode](dem)[minimum height = 1cm, text width=1.85cm][text centered][right=0.25cm of rf]{Demodulator\\(pred)};

	\node[greennode](nn)[minimum height = 1cm, text width=1.25cm][text centered][below=0.25cm of channel]{Neural\\Network};
	\node[greennode](ml)[minimum height = 1cm][text centered][below=0.25cm of rf]{ML};
	
	\draw[->] (mod.east) -- (tf.west);
	\draw[->] (tf.east) -- (channel.west);
	\draw[->] (channel.east) -- (rf.west);
	\draw[->] (rf.east) -- (dem.west);
	\draw[->] (channel.south) -- (nn.north);
	\draw[->] (rf.south) -- (ml.north);

\end{tikzpicture}
\caption{Simplified diagram of the communication system. The ML function takes the signal symbols as its input instead of the received signal itself.}
\label{fig:comm_system}
\end{figure}

The communication system that characterizes the AMC task is shown in Figure \ref{fig:comm_system} with the gray blocks. It is composed of the transmitter, the communication channel, and the receptor. The transmitter is composed of a modulator and a communication filter $r'(x)$. The modulator uses a specific mapping to convert the data into a sequence of symbols. The symbols are upsampled and passed through the communication filter (e.g., root-raised cosine) to avoid inter-symbol interference (ISI). The signal is transmitted through the communication channel, which inevitably adds noise and distortion to the signal. The receptor is composed of another communication filter $r(x)$, that downsamples and transforms the signal back into symbols, and the demodulator, which converts the symbols into data. 

Automatically choosing the correct demodulator based on the signal received is the problem that AMC tries to solve. 
In practice, we distinguish between two scenarios: the coherent scenario, where we have complete knowledge of the system and it is only valid in offline settings; and the non-coherent scenario, where some or all the channel and transmitter characteristics are unknown. The latter is generally used in online settings, as well as to test the model effectiveness in real conditions.

When training neural networks to perform well in AMC or any other task, one optimizes them to be as confident and accurate as possible on a given training set. In the case of AMC, that set is composed of the received signals and their corresponding labels describing the modulation used in the form of one hot encoded vectors. Our claim is that these labels are not good enough to train the model in the AMC task. The reason is that for noisier and heavily distorted signals, there is a non-negligible probability that the signal received was generated from a different modulation than the one we are enforcing the model to predict. Such cases are numerous when testing models in the AMC literature. Therefore, we want train our model with better information so that it can express this uncertainty with less confident predictions. Luckily, in AMC, there is an analytical model that can give us the probability vectors needed to train the model: the Maximum likelihood function.

\section{ML distillation for improved accuracy}


\subsection{Maximum likelihood function}

In many tasks, it can be difficult or even impossible to estimate the true probability of a sample belonging to a particular class. However, in AMC, we can use the Maximum likelihood (ML) function to derive these probabilities. Due to the characteristics of ML, this is only computationally feasible in coherent scenarios, where we have prior knowledge of the channel conditions. Particularly, if we assume that the channel is gaussian with variance $\sigma^{2}$, and that we know the proper communication filter $r(\cdot)$ that matches the filter $r'(\cdot)$ used in transmission, then the ML function is given by the following expression
\begin{equation}
	f_{i}(r(x)) = \dfrac{1}{2\pi \sigma^{2}|M_i|}\prod_{t=1}^{T} \sum_{j=1}^{|M_i|} \exp\left(-\dfrac{\lVert r(x)_t - s_j\rVert^{2}}{2 \sigma^{2}}\right)
\label{eq:ml}
\end{equation}
where $f_{i}$ is the ML function for a given modulation $M_i$ with $|M_i|$ number of states, $T$ is the number of symbols, $x$ is the received signal before the communication filter, $r(x)_t$ is the t-th symbol received, and $s_j$ is the symbol that corresponds to the state $j$. Given a set of modulations, the probability of a modulation $M_i$ would be $p_{ML}^{(i)}(x) = \log(f_{i}(r(x))) / \sum_i \log(f_{i}(r(x)))$. Note that the ML function directly processes $r(x)$ instead of $x$, as shown in Figure \ref{fig:comm_system}.

The reason why ML is not used in practice is simple: they perform badly in non-coherent scenarios, where the channel or transmission filter parameters are unknown. Moreover, any small deviation from the real channel parameters has a major effect on the ML performance, which discards the possibility of trying to estimate these parameters. Finally, incorporating the parameter uncertainty in the ML formulation makes it much more costly computationally, without being more performant than neural networks.

Based on these limitations, we propose a hybrid approach. We will use a neural network for inference, but we will train it with the probabilities computed offline with ML, to give the model this additional uncertainty information. Because Eq. \eqref{eq:ml} is limited to gaussian channels, we will remove the Rayleigh/Rician component of the channel when computing the ML probabilities. This is feasible when we consider that we can have full knowledge of the channel conditions by generating the training samples ourselves. Moreover, after training the neural network, ML is no longer needed, so that our model can be used in non-coherent scenarios, which correspond to real communication systems.

\subsection{Our proposed distillation}

In the AMC literature, neural networks are generally trained with the standard training (ST) loss
\begin{equation}
    \text{ST} : \dfrac{1}{N}\sum_{i}\Ls(f_{\theta}(x_i), y_i)
\end{equation}
where $N$ is the size of the training set.

However, our proposal is to provide the model with labels that better express the uncertainty of highly corrupted signals. Thus, using similar principles as knowledge distillation, we propose to match the neural network outputs with the output of ML in its corresponding coherent scenario.
\begin{equation}
    \text{ST\_ML} : \dfrac{1}{N}\sum_{i}\Ls(f_{\theta}(x_i), p_{ML}(r(x_i)))
\label{eq:st_ml}
\end{equation}
where $p_{ML}$ gives the class probabilities from applying the Maximum Likelihood function, and $r$ is the function that uses the coupled receiver filter and downsamples the signal. Because $r$ and $p_{ML}$ are functions that depend on information that is not known in non-coherent scenarios, we propose to use this loss function only when training the network, where this information can be known.

The main downside of using this function is that, while the probabilities given by ML are optimal in the context of a coherent scenario, they differ from the real probabilities $p^{*}$ that are conditional to the channel and communication filter distributions, $H$. However, our hypothesis, which we test in Section \ref{sec:experiments}, is that training with $p_{ML}$ is going to be a better match to the optimal probabilities $p^{*}$ than the labels $y$ themselves, which rather enforce full confidence on all samples, no matter how noisy they are. Mathematically, these probabilities are related by the following expression:
\begin{equation}
	p[m|x] =  \sum_i p[m|x,H=h_i] \cdot p[H=h_i|x]
\end{equation}
where $m$ is the vector of all modulations, $h_i$ is one possible choice of channel and transmitter parameters. Thus, $p[m|x] = p^{*}$ and $p_{ML} = p[m|x,H=h_0]$. Based on this formulation, our model will match better $p^{*}$ when the true communication parameters $H=h_0$ can be easily predicted from x ($p[H=h_0|x] \approx 1$) or when the communication parameters do not affect that much the probabilities ($p[m|x,H=h_i] \approx p_{ML}, \; \forall i$).


\section{Robust model construction}


Robust models offer more security against adversarial attacks, but they are also more interpretable, basing their decisions on sensible features for the AMC task, like the quadrature and in phase components of the symbols \cite{maroto2021safeamc}. One of the most straightforward ways to improve the network robustness is to use the adversarial training (AT) loss function
\begin{equation}
    \text{AT} : \dfrac{1}{N}\sum_{i}\Ls(f_{\theta}(x_i'), y_i)
\end{equation}
where $x_i' = x_i + \delta_i^*$ is a crafted adversarial signal that maximizes the cross-entropy loss of the neural network in the neighborhood of $x_i$. Because the adversarial perturbation is dependent on the model weights, it has to be computed at each training iteration, typically using PGD.

Analogously to Eq. \eqref{eq:st_ml}, we propose the following adversarial training loss, in which the original labels are replaced with the ML probabilities:
\begin{equation}
    \text{AT\_ML} : \dfrac{1}{N}\sum_{i}\Ls(f_{\theta}(x_i'), p_{ML}(r(x_i')))
\end{equation}
We highlight that the ML function assumes that the signal is distorted with gaussian noise, which is not accurate since the input was adversarially perturbed. However, during adversarial training the adversarial perturbation typically is very small, so we can make the assumption that the combination of the channel gaussian noise and the adversarial noise is approximately gaussian with energy $\sigma_F^{2} = \sigma^{2} + \varepsilon^{2}$.


There is an additional limitation when we are working with adversarial signals. To compute the ML probabilities using Eq. \eqref{eq:ml}, we have to remove the Rayleigh/Rician components of the signal. This is no longer possible, since the adversarial perturbation is constructed from the received signal, and thus, it has not passed through the communication channel. To overcome this limitation, we first remove the Rayleigh/Rician components of the signal, and then compute the adversarial example given by the ML function. Thus, for training the model on non-gaussian channels, instead of AT\_ML, we propose to use the following loss function
\begin{equation}
    \text{AT\_AML} : \dfrac{1}{N}\sum_{i}\Ls(f_{\theta}(x_i'), p_{ML}(r(g_i)'))
	\label{eq:at_aml}
\end{equation}
where $g_i$ is the received signal after removing the Rayleigh/Rician components, and $r(g_i)'$ is a crafted adversarial signal that maximizes the cross-entropy loss of ML in the neighborhood of $r(g_i)$.


\section{Experiments}
\label{sec:experiments}



We propose to evaluate our proposal on scenarios with different levels of non-coherence in the wireless channel. This way, we can verify if the probabilities generated by ML in a coherent scenario are useful to train the model even if it is tested on varying channel and transmitter conditions. 

For all the experiments, we train the neural network to distinguish between the following digital modulations: BPSK, QPSK, 8/16/32/64/128/256-PSK, PAM4, 16/32/64/128/256-QAM, GFSK, CPFSK, B-FM, DSB-AM, SSB-AM, and OQPSK \cite{Rajendran_Meert_Giustiniano_Lenders_Pollin_2018,Guo_Jiang_Wu_Zhou_2020}. For the last six modulations, the ML formulation cannot be directly applied since they shift the signal frequency, phase or amplitude to represent different states, so we use the original labels $y$ on these cases.
For all settings, we consider that the signal sampling frequency is 200KHz. We generate 234000 signals for training and 26000 for testing. The duration of the signal is set to 1024 samples, for consistency with other similar datasets \cite{OShea_West_2016}. For the network, we use a ResNet-based network architecture \cite{OShea_Roy_Clancy_2018}, which trains fast and has great performance in AMC. We train the network for 100 epochs using SGD optimization with momentum 0.9 and learning rate 0.01, decaying exponentially at a rate of 0.95 per epoch. We use gradient clipping of size 5 and weight decay of 5e-4. When computing the adversarial perturbations for both training and testing the robustness of the network, we use PGD-7 \cite{Madry_Makelov_Schmidt_Tsipras_Vladu_2019} with $\varepsilon=20 \text{dB SPR}$ (signal-to-perturbation ratio).

To better showcase the importance of the additional information incorporated by our proposed methods, we also compare our methodology with two other intermediate methods: ST\_LNR and AT\_LNR. These two label noise reduction (LNR) methods, use the ML probabilities to filter from the training all cases where the true label is incorrect. Thus, for ST\_LNR, we train with the ST loss on the subset $D$ given by the signals $x_i$ that fulfill $\argmax(y_i) = \argmax(p_{ML}(r(x_i)))$; and for AT\_LNR, we train with the AT loss on the subset $D'$ given by the signals $x_i$ that fulfill $\argmax(y_i) = \argmax(p_{ML}(r(x_i')))$.

For the first scenario, we consider a gaussian channel with varying levels of noise ranging from -6 to 18 dB SNR. For the transmitter, we use 8 samples per symbol and a root raised cosine filter with rolloff 0.35. Table \ref{tab:sbasic} shows that the network is much more robust when we remove the label noise, both in the standard and adversarial training scenarios, as we would expect from previous intuitions. However, when using the ML probabilities, we get the biggest increase in performance, which underlines the benefit of learning with this additional uncertainty information.

For the second scenario, we additionally vary the transmitter settings. We consider either 2, 4, 8, or 16 samples per symbol and the rolloff of the filter can range from 0.15 to 0.45. Table \ref{tab:sawgn2p} shows very similar results as the previous case. However, since there are more parameter configurations, the difference between $p_{ML}$ and $p^{*}$ increases, which reflects in the lower increase in performance between LNR and ML approaches.

For the third and final scenario, we use the transmitter settings of the first scenario, but we consider also Rician and Rayleigh channels with AWGN noise. 
Table \ref{tab:sp0c20} shows smaller improvements of our method compared to the two previous experimental cases. While the probabilities of ML are not discerning well the label noise, which shows in the lack of improvement for the LNR methods, they can still help the network as seen in the results of the ML methods. Furthermore, we demonstrate that it is important to compute the ML probabilities of the adversarial signal instead of the natural one by showing that our proposed AT\_AML method performs significantly better that an alternative loss function, AT\_NML, in which we compute the ML probabilities of the uncorrupted signal: $(1/N)\sum_{i}\Ls(f_{\theta}(x_i'), p_{ML}(r(g_i))$

To sum up the results obtained over the different scenarios, we show that while removing label noise helps performance, incorporating the uncertainty information with our proposed methods lead to more accurate and robust models.


\begin{table}[htbp]
	\centering
	\begin{tabular}{c|cc}
	    Method & Accuracy & Robustness \\
		\hline
		ST & $85.87 \pm 0.02$ & $19.51 \pm 0.25$ \\ 
		ST\_LNR & $86.34 \pm 0.52$ & $24.61 \pm 1.09$ \\ 
		ST\_ML & $\textbf{86.59} \pm 0.32$ & $\textbf{27.30} \pm 0.47$ \\ 
		\hline
		AT & $57.36 \pm 3.12$ & $54.71 \pm 2.67$ \\ 
		AT\_LNR & $64.50 \pm 0.62$ & $58.44 \pm 0.60$ \\ 
		AT\_ML & $\textbf{67.01} \pm 1.04$ & $\textbf{59.44} \pm 0.32$ \\ 
    \end{tabular}
    \caption{Accuracy and robustness for the ResNet model when only varying the noise settings on a gaussian channel.}
    \label{tab:sbasic}
\end{table}

\begin{table}[htbp]
	\centering
	\begin{tabular}{c|cc}
	    Method & Accuracy & Robustness \\
		\hline
		ST & $85.78 \pm 0.10$ & $20.56 \pm 1.14$ \\ 
		ST\_LNR & $86.21 \pm 0.08$ & $22.82 \pm 0.88$ \\ 
		ST\_ML & $\textbf{86.58} \pm 0.22$ & $\textbf{24.08} \pm 2.66$ \\
        \hline
		AT & $59.61 \pm 6.44$ & $55.45 \pm 3.64$ \\ 
		AT\_LNR & $65.03 \pm 1.23$ & $59.20 \pm 1.34$ \\ 
		AT\_ML & $\textbf{66.27} \pm 1.72$ & $\textbf{59.74} \pm 1.29$ \\ 
    \end{tabular}
    \caption{Accuracy and robustness for the ResNet model when varying the noise level, the samples per symbol, and the transmitter communication filter on a gaussian channel.}
    \label{tab:sawgn2p}
\end{table}

\begin{table}[htbp]
	\centering
	\begin{tabular}{c|cc}
	    Method & Accuracy & Robustness \\
		\hline
		ST & $78.34 \pm 0.02$ & $18.86 \pm 1.83$ \\ 
		ST\_LNR & $78.08 \pm 0.04$ & $18.97 \pm 3.03$ \\ 
		ST\_ML & $\textbf{78.40} \pm 0.22$ & $\textbf{20.00} \pm 1.61$ \\
		\hline
		AT & $50.74 \pm 2.46$ & $\textbf{47.15} \pm 2.53$ \\ 
		AT\_LNR & $48.49 \pm 0.46$ & $45.62 \pm 0.57$ \\ 
		AT\_NML & $49.01 \pm 0.13$ & $45.07 \pm 0.56$ \\  
		AT\_AML & $\textbf{54.10} \pm 0.57$ & $\textbf{46.94} \pm 0.17$ \\ 
    \end{tabular}
    \caption{Accuracy and robustness for the ResNet model when varying the channel properties and the level noise.}
    \label{tab:sp0c20}
\end{table}

\section{Conclusion}

We highlight in this work the benefits of incorporating additional information provided by theoretical models like maximum likelihood when training neural networks. We believe that this is a crucial step to create networks that are more robust, which ensures higher trust in critical scenarios. We showed that networks are more robust when distilling the probabilities given by these mathematical models, even when they operate under different priors.

For future work, it could be interesting to explore other knowledge distillation losses and to use better ML functions that are adapted to Rayleigh or Rician channels, to improve the results in general cases that consider multiple types of channels.

\bibliographystyle{IEEEbib}
\bibliography{references}

\end{document}